\documentclass{article}

\usepackage[preprint,nonatbib]{neurips_2019}

\usepackage[utf8]{inputenc} % allow utf-8 input
\usepackage[T1]{fontenc}    % use 8-bit T1 fonts
\usepackage{hyperref}       % hyperlinks
\usepackage{url}            % simple URL typesetting
\usepackage{booktabs}       % professional-quality tables
\usepackage{amsfonts}       % blackboard math symbols
\usepackage{nicefrac}       % compact symbols for 1/2, etc.
\usepackage{microtype}      % microtypography
\usepackage{graphicx} %package to manage images
\graphicspath{ {./images/} }
\usepackage[labelfont=bf]{caption}
\title{Creative GANs for generating poems, lyrics, and metaphors}

\author{
  Asir Saeed\\
  MLT Labs \\
  \texttt{asir@mltokyo.ai} \\
  \And
  Suzana Ilić \\
  University of Innsbruck, MLT Labs \\
  \texttt{suzana@mltokyo.ai} \\
  \And
  Eva Zangerle\\
  University of Innsbruck \\
  \texttt{eva.zangerle@uibk.ac.at} \\
}


\begin{document}
\maketitle

\begin{abstract}
Generative models for text have substantially contributed to tasks like machine translation and language modeling, using maximum likelihood optimization (MLE). However, for creative text generation, where multiple outputs are possible and originality and uniqueness are encouraged, MLE falls short. Methods optimized for MLE lead to outputs that can be generic, repetitive and incoherent. In this work, we use a Generative Adversarial Network framework to alleviate this problem. We evaluate our framework on poetry, lyrics and metaphor datasets, each with widely different characteristics, and report better performance of our objective function over other generative models. % across architectures.
\end{abstract}

\section{Introduction and related work}
Language models can be optimized to recognize syntax and semantics with great accuracy~\cite{radford2019language}. However, the output generated can be repetitive and generic leading to monotonous or uninteresting responses (e.g ``I don't know'') regardless of the input~\cite{li2015diversity}. While application of attention~\cite{Bahdanau2014,vaswani2017attention} and advanced decoding mechanisms like beam search and variation sampling~\cite{holtzman2019curious} have shown improvements, it does not solve the underlying problem. In creative text generation, the objective is not strongly bound to the ground truth---instead the objective is to generate diverse, unique or original samples. We attempt to do this through a discriminator which can give feedback to the generative model through a cost function that encourages sampling of creative tokens. The contributions of this paper are in the usage of a GAN framework to generate creative pieces of writing. Our experiments suggest that generative text models, while very good at encapsulating semantic, syntactic and domain information, perform better with external feedback from a discriminator for fine-tuning objectiveless decoding tasks like that of creative text. We show this by evaluating our model on three very different creative datasets containing poetry, metaphors and lyrics.

%\section{Related Work}
Previous work on handling the shortcomings of MLE include length-normalizing sentence probability~\cite{Wu2016}, future cost estimation~\cite{Schmaltz2016}, diversity-boosting objective function~\cite{Shao2017,li2015diversity} or penalizing repeating tokens~\cite{DBLP:journals/corr/PaulusXS17}. When it comes to poetry generation using generative text models, Zhang and Lapata~\cite{Zhang2014}, Yi et al.~\cite{Yi2016} and Wang et al.~\cite{Wang2016} use language modeling to generate Chinese poems. However, none of these methods provide feedback on the quality of the generated sample and hence, do not address the qualitative objective required for creative decoding. For the task of text generation, MaskGAN~\cite{Fedus2018} uses a Reinforcement Learning signal from the discriminator, FMD-GAN~\cite{Chen2018} uses an optimal transport mechanism as an objective function. GumbelGAN~\cite{jang2016categorical} uses Gumbel-Softmax distribution that replaces the non-differentiable sample from a categorical distribution with a differentiable sample to propagate stronger gradients. Li et al.~\cite{li2015diversity} use a discriminator for a diversity promoting objective. Yu et al.~\cite{Yu2017} use SeqGAN to generate poetry and comment on the performance of SeqGAN over MLE in human evaluations, encouraging our study of GANs for creative text generation. However, these studies do not focus solely on creative text.

\section{GANs for creative text generation}

Using GANs, we can train generative models in a two-player game setting between a discriminator and a generator, where the discriminator (a binary classifier) learns to distinguish between real and fake data samples and the generator tries to fool the discriminator by generating authentic and high quality output~\cite{Goodfellow2014}. GANs have shown to be successful in image generation tasks~\cite{Denton2015} and recently, some progress has been observed in text generation~\cite{Chen2018, Fedus2018,Yu2017}. Our generator is a language model trained using backpropagation through time~\cite{mozer1995focused}. During the pre-training phase we optimize for MLE and during the GAN training phase, we optimize on the creativity reward from the discriminator. The discriminator's encoder has the same architecture as the generator encoder module with the addition of a pooled decoder layer. The decoder contains $3$ $[Dense Batch Normalization,ReLU]$ blocks and an addtional $Sigmoid$ layer. The discriminator decoder takes the hidden state at the last time step of a sequence concatenated with both the max-pooled and mean-pooled representation of the hidden states~\cite{Howard2018} and outputs a number in the range $[0,1]$. The difficulty of using GANs in text generation comes from the discrete nature of text, making the model non-differentiable hence, we update parameters for the generator model with policy gradients as described in Yu~\cite{Yu2017}. 

%\section{Experimental Setup}
We utilize AWD-LSTM~\cite{merity2017regularizing} and TransformerXL~\cite{dai2019transformer} based language models. For model hyperparameters please to refer to Supplementary Section Table~\ref{tbl:hyper}. We use Adam optimizer~\cite{kingma2014adam} with $\beta1= 0.7$ and $\beta2= 0.8$ similar to ~\cite{Howard2018} and use a batch size of 50. Other practices for LM training were the same as \cite{dai2019transformer} and \cite{merity2017regularizing} for Transformer-XL and AWD-LSTM respectively. We refer to our proposed GAN as Creative-GAN and compare it to a baseline (a language model equivalent to our pre-trained generator) and a GumbelGAN model~\cite{jang2016categorical} across all proposed datasets. We use three creative English datasets with distinct linguistic characteristics: (1) A corpus of $740$ classical and contemporary English poems, (2) a corpus of $14950$ metaphor sentences retrieved from a metaphor database website~\footnote{http://metaphors.iath.virginia.edu/} and (3) a corpus of $1500$ song lyrics ranging across genres. The mix of linguistic styles within this corpus offers the potential for interesting variation during the generation phase. We use the same pre-processing as in earlier work~\cite{Howard2018,howard2018fastai}. We reserve 10\% of our data for test set and another 10\% for our validation set.

%For all datasets, minimal data pre-processing was performed. We forgo lemmatizing, stemming or other morphological modifications to preserve variation.

%\subsection{Training}
We first pre-train our generator on the Gutenberg dataset~\cite{lahiri:2014:SRW} for $20$ epochs and then fine-tune~\cite{Howard2018} them to our target datasets with a language modeling objective. The discriminator's encoder is initialized to the same weights as our fine-tuned language model. Once we have our fine-tuned encoders for each target dataset, we train in an adversarial manner. The discriminator objective here is to score the quality of the creative text. The discriminator is trained for $3$ iterations for every iteration of the generator, a practice seen in previous work~\cite{Arjovsky2017}. Creative-GAN relies on using the reward from the discriminator~\cite{Fedus2018,Yu2017} for backpropagation. We follow a similar training procedure for GumbelGAN. Outputs are generated through sampling over a multinomial distribution for all methods, instead of $argmax$ on the log-likelihood probabilities, as sampling has shown to produce better output quality~\cite{holtzman2019curious}. Please refer to Supplementary Section Table~\ref{tbl:trn} for training parameters of each dataset and Table ~\ref{tbl:hyper} for hyperparameters of each encoder. We pick these values after experimentation with our validation set. Training and output generation code can be found online\footnote{https://github.com/Machine-Learning-Tokyo/Poetry-GAN}.

\section{Evaluation and conclusion}
Evaluating creative generation tasks is both critical and complex~\cite{potash2018evaluating}. Along the lines of previous research on evaluating text generation tasks~\cite{potash2018evaluating}, we report the perplexity scores of our test set on the evaluated models in the Supplementary Section, Table~\ref{tbl:ppx} Our model shows improvements over baseline and GumbelGAN. Common computational methods like BLEU~\cite{Papineni2002} and perplexity are at best a heuristic and not strong indicators of good performance in text generation models~\cite{Theis2016}. Particularly, since these scores use target sequences as a reference, it has the same pitfalls as relying on MLE. The advantages in this approach lie in the discriminator's ability to influence the generator to explore other possibilities. Sample outputs for our model can be found on our website~\footnote{https://www.ai-fragments.com}.
%As reported in Fedus~\shortcite{Fedus2018}, we also notice increases in perplexity in our GAN training phase. 

\iffalse
\section{Discussion}
Since our findings are based on computational metrics, it is difficult to claim that our model manages to understand creativity on a disciplined level. A study with domain experts and with metrics in line with the linguistic definitions of creativity could lead to additional insights.
\fi

%We proposed a GAN learning framework for the generation of creative text,  overcoming some common limitations of text generative models optimizing for MLE. Our framework learns a decoding objective suitable for generation through a learned combination of sub-models that capture creativity and language understanding. Computational metrics shows that the influence of the change in the discriminator's objective shows higher tendency to generate creative output.

\bibliographystyle{unsrt}
\bibliography{neurips}
\newpage
\section{Supplementary Material}
In this section, we report our results on computational metrics, hyperparameters and training configurations for our models. Table~\ref{tbl:ppx} shows the results of the perplexity score evaluation of the evaluated models, Table~\ref{tbl:hyper} shows hyperparameters for each encoding method and Table~\ref{tbl:trn} shows our training parameters. In Table~\ref{tbl:trn}, the values for Gutenberg dataset in columns, GumbelGAN and Creative-GAN are empty as we only pretrain our LMs with the Gutenberg dataset

\begin{table}[h]
\centering
\scalebox{0.8}{
\begin{tabular}{c|ccc|ccc}
\toprule
&
\multicolumn{3}{c|}{\textbf{AWD-LSTM}} & \multicolumn{3}{c}{\textbf{Transformer-XL}} \\ 
 & \textbf{Poetry} & \textbf{Metaphor} & \textbf{Lyrics} & \textbf{Poetry} & \textbf{Metaphor} & \textbf{Lyrics} \\ 
 \midrule
\textbf{LM} & 50.73 & 63.59 & 20.08 & 47.46 & \textbf{62.76} & 16.11 \\
\textbf{GumbelGAN} & 55.03 & 68.72 & 22.19 & 46.27 & 63.43 & 12.58\\ 
\textbf{Creative-GAN} & \textbf{49.40} & \textbf{51.84} & \textbf{17.11} & \textbf{42.45} & 65.35 & \textbf{9.02}\\ 
\bottomrule
\end{tabular}}
\caption{Perplexity Scores, \textbf{bold} denotes best performance}
\label{tbl:ppx}
\end{table}

\begin{table}[h]
\centering
\scalebox{1}{
\begin{tabular}{c|c|c|c|c}%{l|l|l|l|l}
%\begin{tabular}{|l|l|l|l|l}
\toprule
\textbf{Model} & \textbf{W. Emb. Size} & \textbf{Layers} & \textbf{Hidden} & \textbf{Backprop though time}~\cite{mozer1995focused}\\
\midrule
\textbf{AWD-LSTM} & 400 & 3 & 1150 & 70\\
\textbf{Transformer-XL} & 410 & 12 & 2100 & 150\\
\bottomrule
\end{tabular}}
\caption{Encoder Hyperparameters}
\label{tbl:hyper}
\end{table}

\begin{table}[h]
\centering
\scalebox{1}{
\begin{tabular}{c|cc|cc|cc}
\toprule
&
\multicolumn{2}{c|}{\textbf{LM}} & \multicolumn{2}{c|}{\textbf{GumbelGAN}} &
\multicolumn{2}{c}{\textbf{Creative-GAN}}\\ 
 & Epochs & LR & Epochs & LR & Epochs & LR \\
 \midrule
\textbf{Poems} & 8 & $3e-3$ & 10 & $3e-4$ & 10 & $3e-4$ \\
\textbf{Metaphors} & 8 & $3e-4$ & 10 & $3e-4$ & 10 & $3e-4$ \\ 
\textbf{Lyrics} & 15 & $3e-4$ & 12 & $3e-4$ & 12 & $3e-4$ \\
\textbf{Gutenberg} & 20 & $3e-3$ & -- & -- & -- & -- \\
\bottomrule
\end{tabular}}
\caption{Training Parameters}
\label{tbl:trn}
\end{table}

\iffalse
\begin{table}[h]
\centering
\scalebox{0.8}{
\begin{tabular}{l|lll|lll}
\toprule
& 
\multicolumn{3}{c|}{\textbf{AWD-LSTM}} & \multicolumn{3}{c}{\textbf{Transformer-XL}} \\ 
 & \textbf{Poetry} & \textbf{Metaphor} & \textbf{Lyrics} & \textbf{Poetry} & \textbf{Metaphor} & \textbf{Lyrics} \\ 
 \midrule
\textbf{Ground Truth} & 66.21 & 7.49 & 33.32 &  &  &  \\
\textbf{LM} & 79.13 & 7.36 & 44.51 & 37.80 & 63.64 & 43.65 \\
\textbf{GumbelGAN} & 76.79 & 7.32 & 39.58 & 46.27 & 71.57 & 34.55\\ 
\textbf{Creative-GAN} & 75.94 & 7.45 & 39.34 &  &  65.35 & 33.8\\ 
\bottomrule
\end{tabular}}
\caption{DIST Scores}
\label{tbl:dist}
\end{table}
\fi

\end{document}